# SPARSE MODELLING FOR FEATURE LEARNING IN HIGH DIMENSIONAL DATA

[1]HARISH NEELAM, [2]KOUSHIK SAI VEERELLA, [3]SOURADIP BISWAS

[1]PhD student in Biostatistics, Michigan State University, East Lansing, Michigan, USA
[2]PhD student in ECE, Michigan State University, East Lansing, Michigan, USA
[3]MS Data Science, Michigan State University, East Lansing, Michigan, USA
Email: [1]neelamha@msu.edu, [2]veerella@msu.edu, [2]biswasso@msu.edu

**Abstract** - This paper presents an innovative approach to dimensionality reduction and feature extraction in high-dimensional datasets, with a specific application focus on wood surface defect detection. The proposed framework integrates sparse modeling techniques, particularly Lasso and proximal gradient methods, into a comprehensive pipeline for efficient and interpretable feature selection. Leveraging pre-trained models such as VGG19 and incorporating anomaly detection methods like Isolation Forest and Local Outlier Factor, our methodology addresses the challenge of extracting meaningful features from complex datasets. Evaluation metrics such as accuracy and F1 score, alongside visualizations, are employed to assess the performance of the sparse modeling techniques. Through this work, we aim to advance the understanding and application of sparse modeling in machine learning, particularly in the context of wood surface defect detection.

## I. INTRODUCTION

Wood surface defect detection is integral to ensuring the quality of wooden products across various industries. Recent advancements in machine learning have demonstrated promising results in automating this process. Our research aims to contribute to this field by presenting a comprehensive framework that integrates diverse machine learning methodologies, with a particular focus on leveraging sparse modeling techniques.This is crucial for industries such as woodworking and quality control, where early defect detection is paramount for maintaining product quality.

Sparse modeling techniques, notably Lasso and proximal gradient methods, are employed to tackle the challenges posed by high-dimensional datasets. These techniques not only streamline computational processes but also augment the interpretability of extracted features, offering valuable insights into defect characteristics. Our emphasis on sparse modeling extends beyond wood surface defect detection, potentially shaping the development of sparse modeling techniques across various machine learning applications. Our objectives encompass integrating a variety of machine learning techniques to effectively identify wood surface defects while navigating the complexities of high-dimensional datasets through sparse modeling. Moreover, we strive to underscore the importance of interpretability by employing sparse modeling techniques to render results more comprehensible and actionable for industries engaged in wood processing and manufacturing. Ultimately, our research aims to showcase how enhanced defect detection capabilities can translate into tangible benefits such as cost savings and quality enhancement.

## II. RELATED WORK

Feature selection process is a crucial preliminary step in handling high-dimensional datasets. This process aims to reduce dimensionality by selecting a subset of features that effectively capture the distinctions among features concerning the type of label. Achieving feature selection offers numerous advantages, including a better understanding of data with fewer informative features, reduced model complexity and computation time, and the elimination of noisy features.

[1] R. Muthukrishnan and R. Rohini in their published paper, LASSO: A feature selection technique in predictive modeling for machine learning, explored the features of the popular regression methods, OLS regression, ridge regression and the LASSO regression. The performance of these procedures has been studied in terms of model fitting and prediction accuracy using real data and observed promising results.

[2] Maryam A. Alghamdi, Mohammad Ali Alghamdi, Naseer Shahzad, Hong-Kun Xu discussed regarding the iterative methods for solving the lasso which include the proximal-gradient algorithm and the projection-gradient algorithm in their article, Properties, and Iterative Methods for the Lasso.

In our paper, we recognize the significance of addressing these challenges in feature selection, especially in the context of wood surface defect detection. We aim to explore and potentially extend existing methods, incorporating approaches that consider correlated features while balancing computational efficiency and addressing class imbalance concerns. Our focus is on developing a feature selection strategy that aligns with the unique characteristics of our dataset and enhances the interpretability and performance of our defect detection models.

## III. METHODOLOGY

### a. Data

The dataset we used is Wood Defection dataset [3]. It contains 4000 images with annotation for wood surface defects of different types. Some of the examples include No Defect and defect (Quartzity, Live knot, Marrow resin, Dead knot, knot with crack, knot missing and Crack). The original dataset had high resolution images captured with special camera which takes up to 12MB of disk space per image. The size of the dataset has been reduced by resizing the images to 256*256*3 (196608 features). The dataset contains YOLOv5 annotations, which contains the bounding boxes and respective labels as shown in the figure 3.1a.

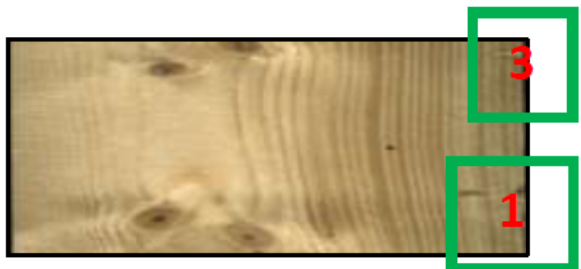

**Figure 1. Image with Bounding Boxes and Labels.**

To tackle the imbalanced nature of our dataset, we framed the task as a binary classification problem focused on detecting defects in wood surfaces. The dataset's structure relies on bounding boxes, where Class 0 signifies defect-free surfaces, and Class 1 denotes defective ones. However, Class 1 instances are approximately one-tenth the number of Class 0 instances, posing a significant challenge due to potential bias towards the majority class.

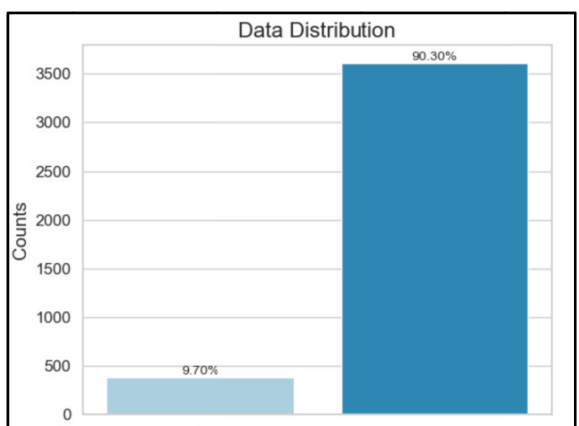

**Figure 2. Data Distribution in Original Data.**

To mitigate this, we utilized Random Over-Sampling, duplicating Class 0 instances to balance representation during training. This strategic augmentation aimed to foster a more equitable distribution, enhancing the model's ability to accurately identify both defective and non-defective wood surfaces and ensuring robustness in classification.

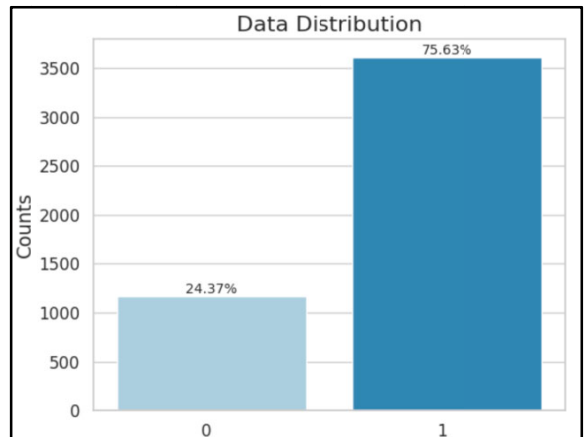

**Figure 3. Data Distribution in Oversampled Data.**

Data preprocessing included several key steps to ensure optimal model performance. Images were resized to 256x256x3 using Lanczos interpolation, a method chosen for its ability to preserve image quality and sharpness. We also converted the images to the RGB color space, which is well-suited for display on screens and has shown to train effectively. For normalization, each image was standardized by subtracting the mean and dividing by the standard deviation of the training set, with these values saved for use during prediction. Additionally, categorical target variables were converted to numerical values using one-hot encoding, transforming them into binary vectors that facilitate the training of our machine learning model.

### b. Feature representation and reduction

In our approach, we integrated several renowned pre-trained models to enhance feature representation and model performance. ResNet50, a deep neural network known for its capacity to efficiently capture intricate image features, formed the basis of our architecture, facilitating the detection of subtle details in wood surfaces. Additionally, we employed the AlexNet architecture, which is celebrated for its pioneering design in computer vision, enabling effective pattern recognition and defect detection on wood surfaces.

Further, we utilized EfficientNetB7 for its efficiency and scalability, which strikes a balance between computational resources and accuracy in capturing detailed patterns from wood surfaces. VGG19 also played a crucial role in our approach; its deep architecture and robust feature extraction capabilities made it the top performer, leading us to select it as our preferred model. This model, trained on extensive datasets like ImageNet, allowed us to fine-tune our approach for specific tasks, such as defect detection in wood, by using transfer learning to handle the challenges posed by our imbalanced dataset.

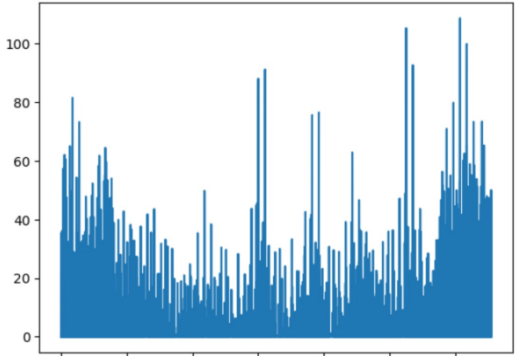
**Figure 4. Feature Representation of an Image by VGG19.**

We then applied feature reduction techniques to evaluate performance enhancements.

Principal Component Analysis (PCA) was used to streamline the feature space. This linear dimensionality reduction technique identified and preserved essential features by projecting data onto orthogonal axes, or principal components. While attempting to balance computational simplicity and information retention necessary for accurate defect detection, the highest accuracy achieved using PCA configurations suggested its limited effectiveness for our specific dataset, with an accuracy of 85.35% and an F-1 score of 0.8594.

To address non-linear relationships in the data, we implemented Kernel Principal Component Analysis (Kernel PCA), which utilizes kernel functions to map data into a higher-dimensional space, aiding in the extraction of complex patterns. Despite efforts to optimize kernel parameters, the top accuracy reached with Kernel PCA also indicated limited success, recording an accuracy of 85.96% and an F-1 score of 0.8646.

The modest performance of both PCA and Kernel PCA could be attributed to several factors: potential loss of crucial information due to dimensionality reduction, increased risk of overfitting where reduced dimensions may not generalize effectively to new data, and the provision of dense solutions that might obscure vital features by including all in the transformed space. These factors collectively could undermine the effectiveness of these techniques in enhancing our defect detection model.

### c. Sparse Modelling Techniques

In our continuous quest to enhance the performance, we explored feature learning techniques with a specific focus on sparse modeling, a pivotal aspect of our goals. Contrary to the expected improvement, feature reduction techniques did not yield the desired enhancements in performance. As a strategic pivot, we shifted our attention to sparse modeling techniques, aiming to selectively emphasize relevant features by inducing sparsity in the feature space. Two prominent sparse modeling techniques were implemented: Lasso regularization and Elastic Net regularization. The rationale behind employing sparse modeling was to accentuate the significance of relevant features while mitigating the impact of irrelevant ones. [2] By designating certain features as sparse (assigned a coefficient of 0), these techniques enabled the identification and prioritization of crucial information for wood surface defect detection. This strategic shift toward sparse modeling reflects our commitment to adapt and refine our approach based on empirical results, ultimately steering the project toward its primary goal of efficient and accurate defect detection in wood surfaces.

### Lasso Regularization:

Lasso regularization serves as a potent tool in preventing overfitting, a common challenge in machine learning models. Its mechanism involves augmenting the standard least square's objective function with a penalty term proportional to the absolute values of the coefficients within the regression model as shown in the Equation (1). This added L1 penalty induces sparsity in the model by driving certain coefficients to exactly zero. In the context of our wood surface defect detection project, we applied Lasso regularization to optimize the cost function, which includes both the least square errors and the L1 penalty term. During the training phase, where the model learns from labeled data, the objective is to minimize this cost function. [4] This process entails determining the coefficients that minimize the combined impact of least square errors and the L1 penalty. These coefficients, once identified through the training process, are then employed to predict features in testing instances. The regularization parameter, denoted as lambda, plays a pivotal role in governing the strength of regularization. A higher value of lambda indicates more potent regularization, effectively emphasizing sparsity in the model. This fine-tuning parameter allows us to strike a balance between fitting the model to the training data and preventing it from becoming overly complex, thereby contributing to robust and effective defect detection in wood surfaces.

$$\text{Cost function} = \frac{1}{2m}\sum_{i=1}^{m}\left(y_{p}\text{red}(i) - y_{t}\text{rue}(i)\right)^{2}$$
$+\lambda \sum_{j=1}^{n}|\text{coef}|$ ---- Equation (1)

The Lasso regularization was implemented using the library from scikit-learn. Setting the regularization constant (Lambda) to 0.01, the application of Lasso regularization resulted in a notable reduction of features for each image. Specifically, the feature count diminished from 32,768 to 2,026, signifying a substantial simplification of the dataset. The efficacy of this reduction lay in its ability to retain relevant information while discarding less significant features. The refined set of 2,026 features was then employed to train and fit a KNN model. This sequential approach not only facilitated computational efficiency but also aimed to enhance the model's performance by focusing on the most informative attributes derived through Lasso regularization.

- Accuracy: 89.12%
- F-1 score: 0.8974.

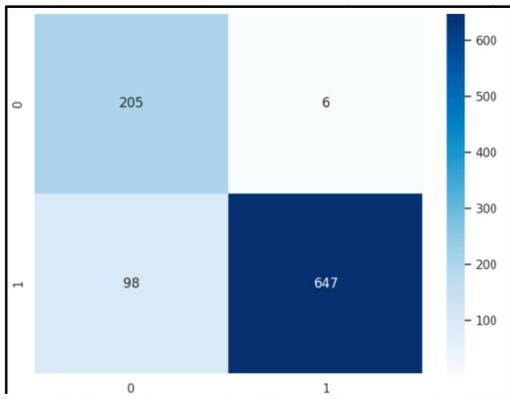

**Figure 5. Confusion matrix of Lasso Features using KNN.**

**Elastic Net Regularization:**
Elastic Net regularization [8], a hybrid approach encompassing both L1 and L2 regularization, offers a nuanced solution by striking a balance between the sparsity-inducing nature of Lasso and the grouping effect of Ridge as shown in equation (2). In our implementation, we set the hyperparameters to an alpha value of 0.01 and an L1_ratio of 0.5, signifying equal importance given to both Lasso and Ridge regularization. Notably, a L1_ratio of 1 emphasizes Lasso, while 0 emphasizes Ridge, allowing us to fine-tune the regularization strategy based on the characteristics of our data. Remarkably, our experimentation with Elastic Net yielded results comparable to those obtained with Lasso in terms of selected features and performance scores. However, the similarity in results suggests that the functionality of Ridge, a prominent component of Elastic Net, might not be particularly suitable for our specific case. Ridge regression is particularly beneficial when dealing with datasets that exhibit highly correlated features. In our context, the dataset might lack such high correlations, potentially rendering the Ridge component less impactful. The adaptive nature of our approach acknowledges the nuances of our dataset, underscoring the importance of tailored regularization strategies to optimize the performance of our wood surface defect detection model.

$$\text{Cost function} = \frac{1}{2m}\sum_{i=1}^{m}(y\_pred(i) - y\_true(i))^2$$
$$+ \alpha\left(\lambda_1 \sum_{j=1}^{n}|coef(j)| + \frac{\lambda_2}{2}\sum_{j=1}^{n}(coef(j))^2\right)$$

---- Equation (2)

**Optimization:**
With a streamlined dataset refined through Lasso regularization, our project pivoted towards optimization strategies to boost efficiency and predictive accuracy. We integrated the proximal gradient optimization technique to fine-tune feature representation, enhancing model performance. This advanced strategy goes beyond traditional methods by adjusting features refined through Lasso regularization, aiming to balance sparsity with accuracy—these aid in retaining essential information while minimizing extraneous details.

We employed the proximal gradient descent method specifically on Lasso features, using the following objective function equation.

Objective Function = $0.5 * \|Ax - b\|_2^2 + \lambda \|x\|_1$ --------Equation (3)

The training targets, with λ set at 0.1. Optimization was facilitated by the L-BFGS-B algorithm from the SciPy library, which is tailored for bound-constrained optimization, ensuring coefficients stay within defined bounds. After 10 optimization cycles, we significantly reduced the feature set from 32,768 to just 1,179, maintaining robust performance with an accuracy of 89.65% and an F-1 score of 0.902.

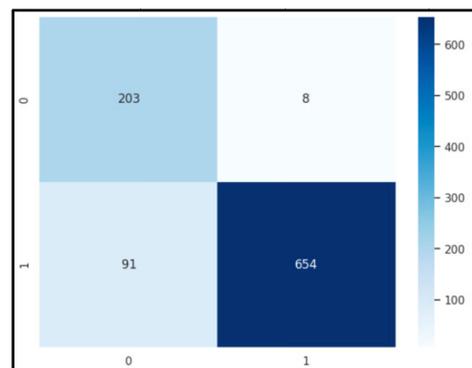

**Figure6. Confusion matrix of Proximal Gradient Features using KNN.**

The combination of Lasso regularization and proximal gradient optimization provides considerable benefits, particularly in handling high-dimensional datasets. These techniques not only enhance feature selection and model interpretability but also increase computational efficiency. By promoting sparsity, where non-essential features are driven to zero, these methods refine the feature set, thereby improving the model's predictive accuracy and ensuring a streamlined, efficient modeling process.

## IV. RESULTS

In addressing the challenges posed by imbalanced data in wood surface defect detection, our investigation focused on applying sparse modeling techniques to refine feature sets and enhance model performance. Initial results from the original imbalanced dataset yielded an accuracy of 84.75% and an F1-score of 0.8376. Subsequent sparsing with Lasso regularization, adjusting the lambda value to 0.01, resulted in a reduced feature set of 1932, showing slightly adjusted accuracy of 84.5% and an F1-score of 0.829. Comparatively, Elastic Net and Proximal Gradient Descent presented slight variations in performance, emphasizing the complexities and subtle nuances of handling severely imbalanced datasets with high-dimensional data.

Further comparative analysis between feature reduction and sparse modeling techniques like PCA and Lasso revealed distinct advantages depending on the data characteristics and the modeling objectives. While PCA reduces dimensionality irrespective of class labels, potentially overlooking crucial classification features, Lasso drives certain coefficients to zero, prioritizing features highly correlated with class outcomes. This inherent prioritization in Lasso and similar techniques supports more robust classification performance, particularly in scenarios requiring effective feature discernment. The results underscore the effectiveness of sparse modeling techniques, such as Lasso and proximal gradient descent, in optimizing machine learning models for defect detection in wood surfaces by enhancing interpretability and predictive accuracy.

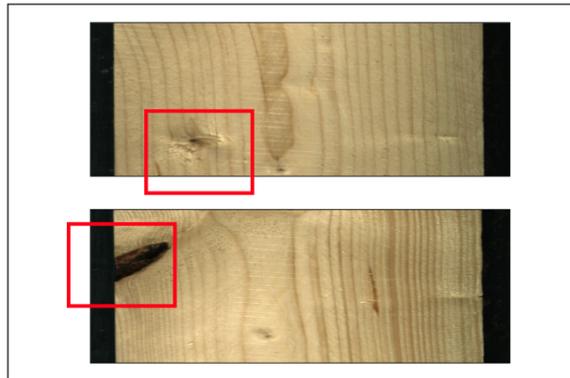

**Figure 7. Images representing the relevant portions of features.**

## V. CONCLUSION

In the exploration of sparse techniques for feature selection and optimization, the efficacy of such methodologies has been evident, with a notable impact on our model. The success of Lasso regularization and Proximal Gradient Descent (PGD) underscores the importance of tailoring feature selection strategies to the characteristics and dimensionality of the dataset at hand. Despite the limited number of features, both Lasso and PGD exhibited commendable performance, showcasing their ability to capture essential information and facilitate accurate classification. The judicious application of these sparse techniques not only streamlined the feature space but also contributed to the overall robustness and interpretability of the model. The results emphasize the significance of choosing techniques that align with the inherent structure of the data, particularly when faced with challenges such as dimensionality reduction and feature relevance.

As we conclude, the success of Lasso and PGD in our model serves as a testament to the nuanced interplay between feature selection methods and dataset characteristics. This underscores the need for a tailored, data-driven approach in selecting techniques that can unlock the full potential of machine learning models, ultimately paving the way for enhanced performance and insightful analyses in the realm of wood surface defect detection.

★★★